# An Empirical Comparison of Explainable Artificial Intelligence Methods for Clinical Data: A Case Study on Traumatic Brain Injury

Amin Nayebi[1], Sindhu Tipirneni[2], Brandon Foreman[3], Chandan K Reddy[2], Vignesh Subbian[1]

[1]The University of Arizona, AZ, USA; [2]Virginia Tech, VA, USA; [3]University of Cincinnati, OH, USA

**Abstract**

*A longstanding challenge surrounding deep learning algorithms is unpacking and understanding how they make their decisions. Explainable Artificial Intelligence (XAI) offers methods to provide explanations of internal functions of algorithms and reasons behind their decisions in ways that are interpretable and understandable to human users. . Numerous XAI approaches have been developed thus far, and a comparative analysis of these strategies seems necessary to discern their relevance to clinical prediction models. To this end, we first implemented two prediction models for short- and long-term outcomes of traumatic brain injury (TBI) utilizing structured tabular as well as time-series physiologic data, respectively. Six different interpretation techniques were used to describe both prediction models at the local and global levels. We then performed a critical analysis of merits and drawbacks of each strategy, highlighting the implications for researchers who are interested in applying these methodologies. The implemented methods were compared to one another in terms of several XAI characteristics such as understandability, fidelity, and stability. Our findings show that SHAP is the most stable with the highest fidelity but falls short of understandability. Anchors, on the other hand, is the most understandable approach, but it is only applicable to tabular data and not time series data.*

## 1 Introduction

Over the past several decades, the area of machine learning and artificial intelligence (AI) has advanced in terms of methods and applications to domains such as biomedicine. The driving forces behind this trend include early methodological contributions in support vector machines, as well as more recent advances in deep learning techniques, particularly those based on self-supervision or weak-supervision [1]. While these models show strong prediction accuracies, their hierarchical non-linear structure renders them opaque, i.e., it is unclear what information in the input data causes them to make certain predictions [2]. Although several studies have attempted to incorporate explainable modules into deep learning architectures [3–5], traditional AI approaches are generally fraught by skepticism from domain experts and the general public, demanding further development of approaches for "opening" black-box models [6] and providing useful and appropriate explanations to different users [7].

The ability to comprehend and validate an AI system's decision-making process and outcomes are an important prerequisite to clinical deployment. Rather than relying on the forecasts of a black box system by default, it may be only appropriate to offer available decision options and the associated explanations to a human user for further evaluation, confirmation, and action. Explainable Artificial Intelligence (XAI) attempts to provide end users with such understandable explanations [8]. XAI approaches also conform to ethical concerns and regulatory considerations that must be addressed when prejudice or discriminatory findings are found [9]. XAI is expected to give much-needed trust and confidence in human-AI cooperation for key applications in healthcare. Given the variety of accessible XAI approaches, deciding which explanation is best suited for a specific application area is a challenge [10].

The objective of this work is to implement and empirically compare six XAI approaches to predictive models using two different forms of clinical data, structured tabular and time series physiologic datasets. To support our demonstration, we focus on acute Traumatic Brain Injury (TBI) as the clinical use case [11]. TBI is one of the primary causes of mortality and disability in the United States, with roughly 2.8 million new TBI cases reported every year [12]. We selected TBI as the use case for this work because the pathophysiology and clinical course of TBI patients tends to be highly variable and extremely challenging to prognosticate without advanced monitoring and analytical techniques. The specific questions we sought to answer in this work are (1) What are the characteristics and practical considerations of XAI techniques when applying them to clinical prediction problems? and (2) How to compare and evaluate XAI techniques in the context of understandability, stability, and fidelity?

## 2 Materials and Methods

### 2.1 Source of Data

We utilized data from the prospective, multicenter Transforming Research and Clinical Knowledge in Traumatic Brain Injury (TRACK-TBI) study [13]. TRACK-TBI collected detailed clinical data on nearly 3000 TBI patients from 18 academic Level I trauma hospitals throughout the United States. For this case study, we used two subsets of data from the TRACK-TBI study: (1) structured clinical data and (2) high-resolution, continuous monitoring physiologic data

Structured data was comprised of clinical variables at the time of admission and during the initial five days of hospital stay across patients admitted to the hospital with TBI and included clinical assessments such as the Glasgow Coma Scale (GCS) score and its motor, verbal, and eye sub-scores; ventilatory support; laboratory values; and vital signs gathered by clinical staff. Patients admitted to intensive care had vital signs and clinical assessments documented hourly for the first five days of care or through discharge, whichever was earliest.

A subset of patients (n = 25) underwent detailed physiologic data capture using a bedside data aggregation device (Moberg Solutions, Inc; Ambler, PA) at a single enrolling TRACK-TBI site. These patients' waveform-level data consists of standard vital sign monitors such as heart rate and arterial blood pressure as well as intracranial monitoring data, such as intracranial pressure where available. Waveform data was acquired at 125 Hz and down sampled to 1 Hz for the purposes of this study.

### 2.2 Prediction Models

We developed two prediction models for each dataset. For the structured clinical dataset, we included 900 out of 2996 participants, who had outcome data and recordings of blood pressure for at least 12 hours in the first 48 hours of ICU stay. Only variables with less than 50% missing data were included. The mean statistic for each variable were used to impute missing values. Each variable was averaged over the first 24 hours of the ICU stay. We then trained a Support Vector Machine (SVM) on the preprocessed data to predict a dichotomized favorable versus unfavorable outcome, 6-month post-injury, as defined by Glasgow Outcome Scale-Extended (GOSE). More details of study selection criteria are described in [11].

A separate deep Recurrent Neural Network (RNN) model was developed for the continuous monitoring physiologic data. We included 16 out of 25 patients for whom the Intracranial Pressure (ICP) was recorded. Physiologic variables with insufficient data were excluded and consequently, nine variables were retained. Each two-hour window was considered a separate sample, which amounts to a total of 2,880 data instances. Missing time points are interpolated linearly. A RNN model using Long Short-Term Memory (LSTM) units was developed for predicting unfavorable results. A sample is labeled with an unfavorable outcome if there is an adverse event in the one hour following the input window. Here, an adverse event is defined when ICP is larger than 22 mmHg for at least 15 minutes.

We used both a deep learning model (RNN) and a more traditional machine learning (SVM) method to show the flexibility of XAI methods on different prediction model structures.

### 2.3 Interpretation Methods

Explainability and interpretability are often used interchangeably in the context of machine learning and AI. Interpretability is considered by some to be a component of explainability; that is, explainable models are interpretable by default [14]. In this study, both the terms - interpretability and explainability - refer to the extent to which a system's internals are understandable by humans. We implemented and analyzed six different XAI methods (see Table 1) in the study. In this section, we briefly introduce and explain each of the techniques.

*Local Interpretable Model-agnostic Explanations (LIME):* The main idea behind LIME is to approximate the behavior of the prediction model in the vicinity of an instance of data [15]. The explanations produced in the vicinity of $x$ is:

$$\xi(x) = \underset{g}{\mathrm{argmin}}\ \mathcal{L}(f, g, \pi_x) + \Omega(g)$$

Here, $\xi(x)$ is the best explanation around sample $x$ based on all explanation functions $g$ (e.g., linear regression model) that minimize the loss function $\mathcal{L}$. $\pi_x$ is a weight function showing how close a data point is to the original $x$, and $\Omega(g)$ is the complexity of function $g$. LIME usually uses a linear regression as the explanation function $g$ in the form of $g(x) = w_g x$. To calculate the contribution of each variable to the outcome, each feature value is multiplied by a corresponding weight parameter. Let $\varphi_i$ be the contribution score of $i^{th}$ feature, then the following equation shows how to derive the importance of each variable for a given instance:

$$g(x) = \sum_{i=1}^{M} w_i x_i = \sum_{i=1}^{M} \varphi_i \rightarrow \varphi_i = w_i x_i$$

*SHapley Additive exPlanation (SHAP):* SHAP [16], on the other hand, does not approximate the prediction function, rather, assigns contribution scores to each feature. However, the definition of Shapley values is very similar to LIME. The explanation function for the Shapley value is $g(x') = \phi_0 + \sum_{i=1}^{M} \phi_i x'_i$ which is the linear function over interpretation representation of each instance $x'$, showing the presence or absence of each feature. SHAP shows that only one possible explanation function $g$ can satisfy three properties of local accuracy, missingness, and consistency.

*Partial Dependence Plot (PDP):* PDP is a simple tool to visualize the effect of one or two features on the outcome of a black-box function, first introduced in [17]. PDP can demonstrate whether the relation between the output and the input is linear, monotonic, or more complex. Partial dependent function is derived as follows:

$$f_S(x_S) = \frac{1}{n}\sum_{i=1}^{n} f(x_S, x_C^{(i)})$$

Here, $x_S$ is the subset of features for which the partial dependence function is desired and $x_C^{(i)}$ represents remaining features in the $i^{th}$ sample. The features in S are those for which we want to know the effect on the prediction. The partial function shows us what the average marginal effect on prediction is for a given value of feature S. This method assumes that the characteristics in C are unrelated to the characteristics in S. If this assumption is broken, the partial dependency plot's averages will include data points that are extremely uncommon, if not impossible.

*Explanation-by-example:* Another XAI approach is to use examples from the training dataset to capture the link between a particular test input and the underlying training data that influenced the model's output. We use an explanation-by-example method implemented by [10] called *ExMatchina*, which uses the training dataset's closest matching data samples as representative examples. By comparing feature activations at one of the architecture layers, the closest instances are chosen. This method is only applicable to deep learning models since it needs the output of one of the last layers of the model. We used the cosine similarity function, which is validated by [18], as a distance metric for finding the closest neighbors.

*Anchors*: The Anchor method clarifies individual predictions made by a prediction model using certain *if-then* rules that "anchor" the prediction [19]. In other words, if the identified rules hold, the prediction most will likely not change, even if the rest of the features are perturbed (i.e., are assigned random values). Using this approach, the most important features that anchor a label on an individual instance are determined. A rule's *precision* is defined as a percentage of neighboring samples of example $x$ that, by holding the rule, will have the same output as $x$. We set the precision to be 95% i.e., Anchors' explanations only contain the rules by holding which the predicted label will not change in 95% of the time in the vicinity of an input instance. The *coverage* of an anchor is the likelihood of rules not being violated in the vicinity of sample $x$.

**Table 1.** List of datasets/prediction models and interpretation methods that were used for each dataset.

| Explanation Type | Explanation Method | Structured Clinical Data (SVM Model) | Physiological Data* (RNN Model) | Level of explanation** |
|---|---|---|---|---|
| Feature relevance | SHAP | ✓ | ✓ | G & L |
|  | LIME | ✓ | ✓ | G & L |
| Rule-based | Anchors | ✓ | ✗ | L |
| Example-based | Counterfactual example | ✓ | ✗ | L |
|  | Explanation-by-example (ExMatchina) | ✗ | ✓ | L |
| Visual explanation | PDP | ✓ | ✗ | G |

\* To run perturbation-based methods on the physiologic data, we considered each time step of each variable as a feature, so the methods could compute the contribution score corresponding to each single time step.
\*\* G stands for global interpretation and L represents the local interpretation

*Counterfactual Explanations:* A counterfactual explanation describes a situation in which "if the event X had not happened, Y would have occurred" [20]. In machine learning, a counterfactual explanation identifies new values for certain features that lead to the new label for prediction. A counterfactual explanation is of quality if it satisfies two conditions. First, the generated counterfactual example should be as similar as possible to the input instance. Second, the counterfactual explanation must change as few features as possible compared to the original input instance. We use the counterfactual method presented by [21] in which the loss function is as follows:

$$L(x, x', y', \lambda) = \lambda \cdot (f(x') - y')^2 + d(x, x')$$

Here, $x'$ is the counterfactual example, $y'$ is the desired counterfactual label, and $d(.)$ is the distance function. The loss function minimizes both (a) the distance between input instance and the counterfactual example and (b) the distance between the predicted outcome for the counterfactual example and its desired label. This implementation of counterfactual examples does not restrict the number of features that will change in the counter example.

The implementation of this study was done using Python 3. The code for this implementation is available on GitHub[1].

### 2.4 Evaluation of XAI Techniques

We compared results at both local and global levels of interpretability. The goal of *local interpretation* is to explain a single prediction by focusing on a particular instance and attempting to comprehend how the model reached at its prediction. *Global interpretation*, on the other hand, refers to the ability to comprehend the distribution of the prediction output based on the input features [22]. We then analyzed the six selected XAI techniques for each of the following attributes.

- *Stability*: The coherence of a XAI technique over the same input is characterized as stability [23].
- *Ease of configuration*: The ease with which a method's hyperparameters may be tuned shows how a method is configurable. Different configurations are one of the factors that affect stability.
- *Fidelity*: It is the closeness of the explanations to the model predictions, showing how well the explanation approximates the black-box model [22].
- *Understandability*: It is the capacity to provide human-understandable explanations [24].

## 3 Results

### 3.1 Interpretation based on Structured Clinical Data

*Local interpretation:* SHAP, LIME, Anchors, and counterfactual example methods were applied on SVM prediction model for local interpretations. We were not able to implement the explanation-by-example, since it was designed to explain only deep learning models. Using each XAI method, we show the results for a patient A with favorable outcome that is predicted by the model as unfavorable, incorrectly. Figure 1 shows the weights of the linear approximation of the prediction function around patient A using the LIME.

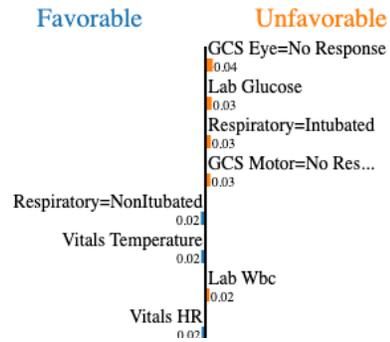

**Figure 1**. Weights of the covariates in the LIME linear regressor which approximates the prediction localy around patient A. Variables under the left column contribute to the unfavorable outcome and vice versa

Figure 2 visualizes the result of local interpretation for patient A. Section (b) of Figure 2 illustrates how each variable is contributing to the outcome based on the adjusted LIME importance scores. To calculate these scores, we multiplied the value of each variable by its corresponding weight in Figure 1. While LIME and SHAP assign relevance scores to each feature, Anchors and counterfactual examples offer other sorts of explanation, namely if-then rules and altered instances. In the counterfactual example (Figure 2-d), almost all the features were changed, but for the sake of interpretability, we only report features with more than 0.5 change in their normalized values (all features are normalized using z-score).

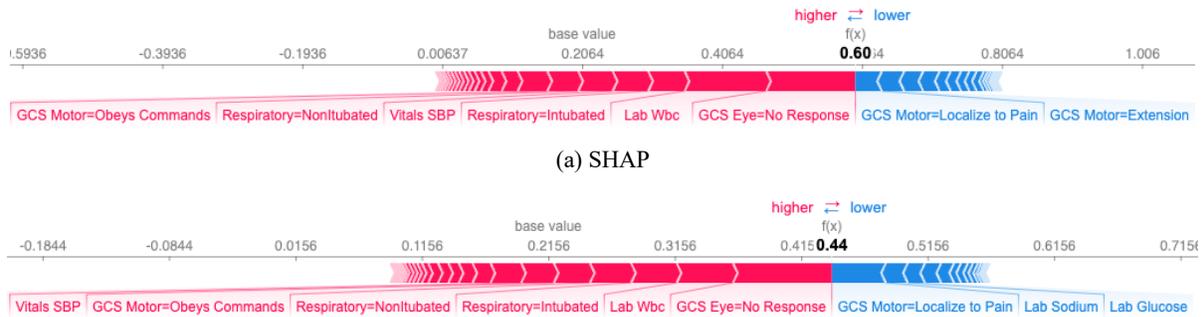

(a) SHAP

[1] https://github.com/aminnayebi/XAIComparison

```
Step 1: (GCS Motor=Untestable) > 0.00
        Precision = 34 percent and Coverage = 6.37 percent
------------------------
Step 2: (GCS Eye=No Response) > 0.45
        Precision = 70 percent and Coverage = 1.38 percent
------------------------
Step 3: (Vitals Temperature) <= 36.73
        Precision = 90 percent and Coverage = 0.32 percent
------------------------
Step 4: (Lab Magnesium) > 1.90
        Precision = 100 percent and Coverage = 0.17 percent
```

```
Original predcited label is: Unfavorable
---------------
Lab Glucose changed from 126.33 to 97.31
New predcited label is: Favorable
---------------
Lab Wbc changed from 22.4 to 19.25
New predcited label is: Favorable
---------------
GCS Eye=No Response changed from 1.0 to 0.68
New predcited label is: Favorable
```

(c) Anchors                                   (d) Counterfactual Example

**Figure 2.** The results of different interpretation methods locally around patient A. In both SHAP and LIME plots, features indicated with red and blue colors contribute to the unfavorable and favorable outcomes, respectively. The reported coverage and precision in Anchors are cumulative, i.e., in each step, the precision and coverage are related to all the rules stated so far.

*Global interpretation:* Out of four explanation methods that were used for local interpretability of structured data, we were able to build global interpretations using only SHAP and LIME. To do so, variables are ranked based on the magnitude of their importance score for the entire test data. We also included the results of PDP which is global explanation method by default. The results are depicted in Figure 3. Explaining the behavior of the prediction model globally using Anchors and counterfactual examples is not trivial and beyond the scope of this work.

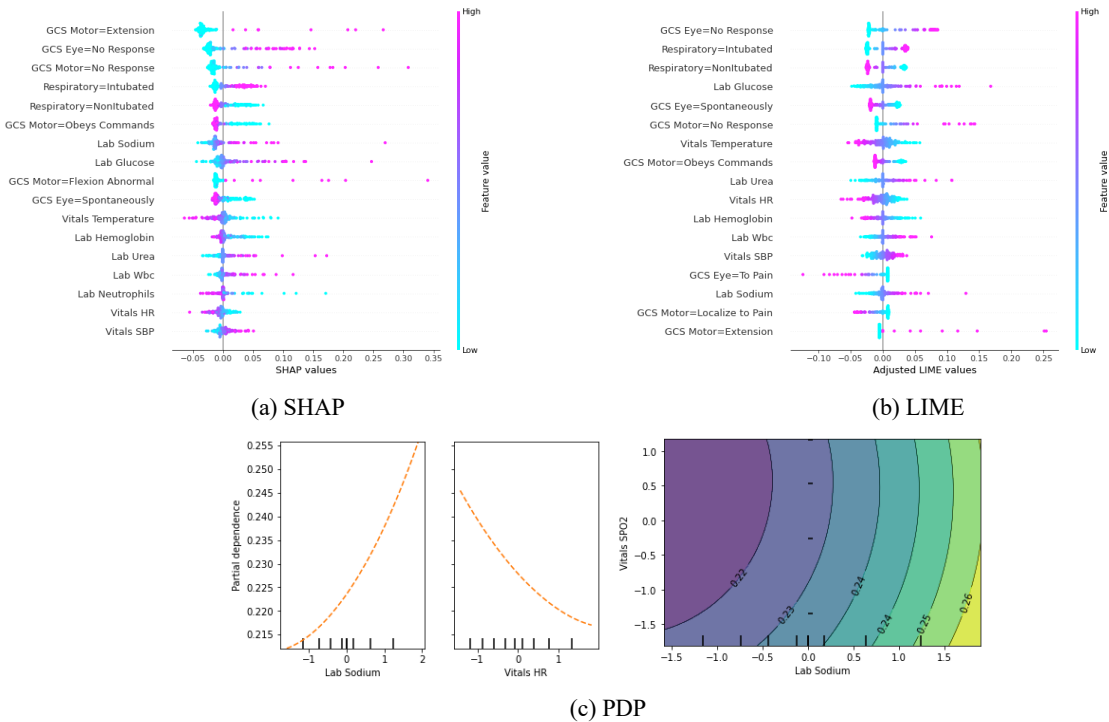

(a) SHAP               (b) LIME

(c) PDP

**Figure 3.** (a) and (b): The most important features based on the absolute value of SHAP/LIME. Each point represents a patient record, and the horizontal axis reflects the SHAP/LIME value of each patient's characteristic. SHAP/LIME levels that are negative or positive indicate that they contributed to a favorable or unfavorable result, respectively. Each point's color symbolizes the value of a feature to a patient. (c): Partial dependence plots for three variables. In two plots on the left, horizontal axis corresponds to the values of a feature and the y-axis represents the output of the prediction model (i.e., the chance of having an unfavorable outcome). The right figure shows the effect of two values on the prediction, simultaneously.

### 3.2 Interpretation based on Physiologic Data

*Local interpretation:* Since physiologic data are time-series data, we could only implement LIME, SHAP, and explanation-by-example (ExMatchina) methods to locally interpret an instance of the physiologic data. Figure 4 and Figure 5 present the local interpretation for a sample with an unfavorable outcome that the model also predicted correctly. Figure 4 illustrates the results from SHAP and LIME methods for top three important variables. Figure 5 shows the results of the explanation-by-example method in which two closest examples from the training dataset are presented.

*Global interpretation:* To explain the model globally, we could only use LIME and SHAP by aggregating the importance scores of each time step and evaluating the contribution score for each variable. From there, the global interpretation plots were made similar to the structured data. Figure 6 shows the top important features. To assign a single value to each temporal feature, we took an average over all data for a sequence.

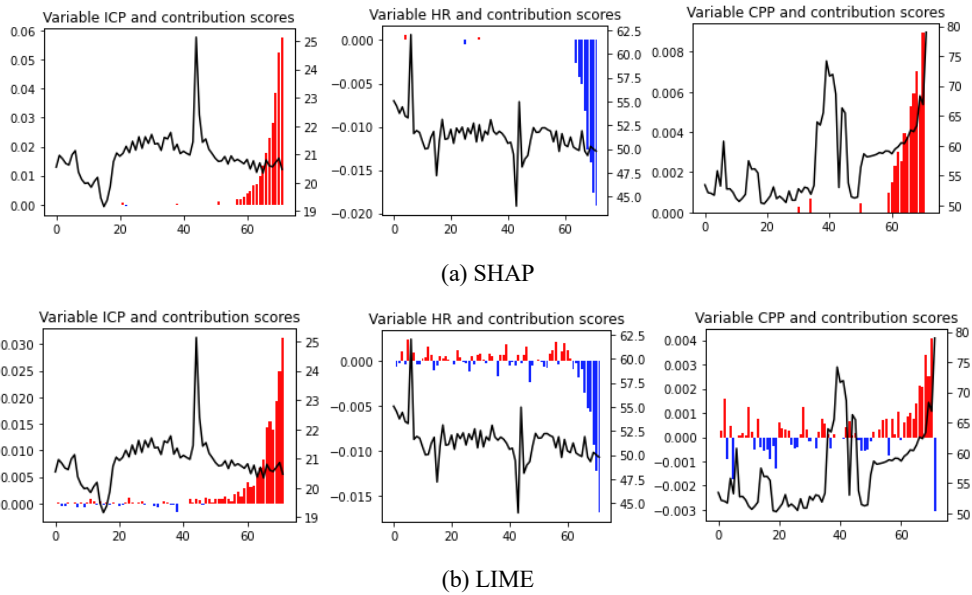

(a) SHAP

(b) LIME

**Figure 4.** Temporal values and their corresponding contribution score (SHAP and LIME) for top three important features of an unfavorable sample. The solid black line illustrates the temporal values and blue and red bars show how much each time steps is contributing to a favorable and unfavorable outcome. *ICP*: Intracranial Pressure, *HR*: Heart Rate, *CPP*: Cerebal Perfusion Pressure

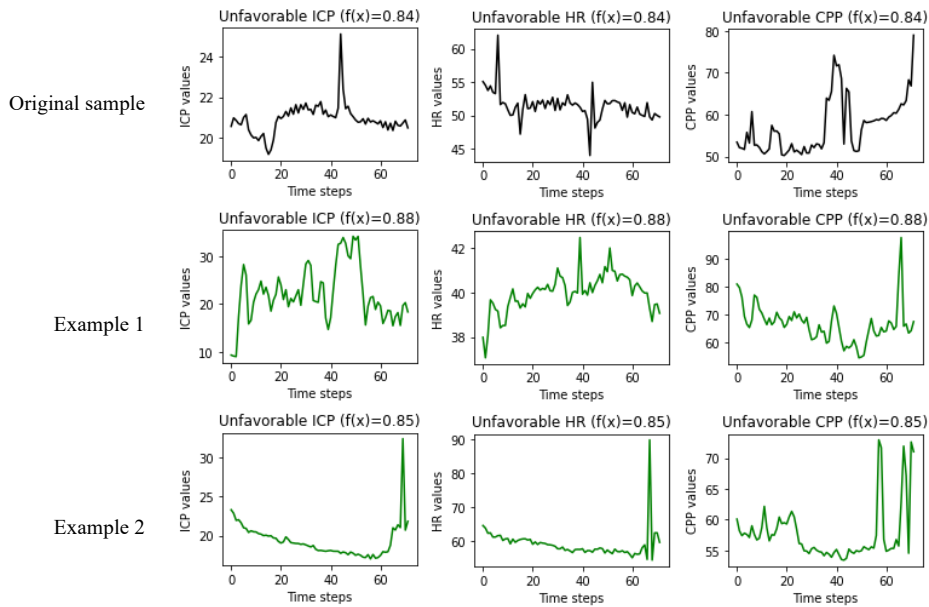

**Figure 5.** The results of the explanation-by-example method for a single instance of physiologic data. The first row, shows the original sample values for the top three important features, and the next two rows are the closest examples from training data. The value of *f(x)* in the title of each plot, shows the output of the prediction model for the corrsponding sample.

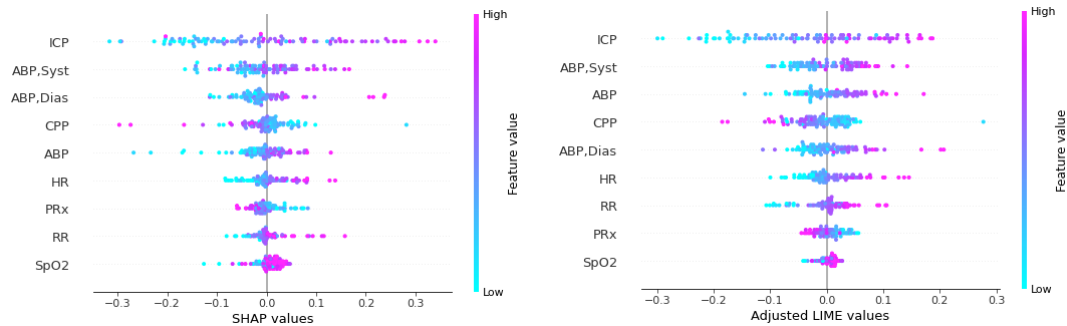

**Figure 6.** The most important features based on the absolute value of SHAP/LIME. Each point represents a patient, and the horizontal axis reflects the SHAP/LIME value of each patient's characteristic. SHAP/LIME levels that are negative or positive indicate that they contributed to a favorable or unfavorable result, respectively.

## 4 Discussion

Ease of configuration, stability, fidelity, and understandability are important attributes for XAI methods in a clinical environment. Ease of configuration is crucial so that the burden of activities such as fine-tuning an explanation algorithm does not fall on end users (e.g., clinicians and patients). The explanations must be stable enough for people to trust the AI system and produce the same outcomes each time they are required. In addition, the consistency of the explanation to the prediction model is essential and may be measured by an explanation's fidelity. Importantly, explanations must be intelligible, particularly in sectors such as healthcare where users may not be familiar with how prediction models and explanations function. In the following subsections, a comparison of XAI methods is provided based on aforementioned attributes.

### 4.1 Comparison of XAI Techniques

LIME generates the weights of each covariate in the approximated linear regression approach at the local explanation level. It is important to note that the weights do not convey much information, particularly when data are normalized. The positive and negative weights do not accurately represent the variables' contributions to the prediction model for a specific instance. For example, LIME assigns a positive weight to glucose for patient A (Figure 1), indicating that higher glucose increases the likelihood of a negative result. It does not imply that glucose is a factor in the unfavorable result in this particular sample (patient A). Rather, we can see (Figure 2-b) that the glucose is contributing to a favorable outcome for patient A by computing the contribution score for the variable.

At the local level, the SHAP and LIME outcomes are extremely similar. Seven of the eight important features based on SHAP for patient A are also among the important variables derived from LIME. However, when comparing the value of timesteps for physiologic data between LIME and SHAP, one may infer that SHAP focuses on fewer time steps, while LIME spreads the importance over the time spectrum. For example, in the CPP sequence (see Figure 4), unlike SHAP, which awards practically all of the scores to the final 12 timesteps, LIME assigns a significant amount of credit to each and every time step. This tendency may also be seen in the global interpretation visualizations for the structured data. Figure 3 displays more cases with higher SHAP contribution scores, indicating that, in comparison to LIME, SHAP pays less attention to the remaining characteristics in those samples. For example, we can observe that the LIME scores in Figure 3 are less dispersed than SHAP scores for the variable "GCS Motor=No response".

The LIME and SHAP scores are relatively simple to work with. For structured data, simply summing the contribution scores across samples for each feature allows one to shift from local to global interpretation. For physiologic data, the contribution scores may be obtained at both time and feature levels by simply aggregating the contribution scores.

The results provided by Anchors are not very consistent with the results of SHAP and LIME for the structured data. From four different rules that Anchors presented, only one of the features (i.e., GCS eye) is among the important variables presented by LIME and SHAP. One of the features (magnesium) is not even among top 17 important variables that are described by global explanation of SHAP and LIME (Figure 3). On the other hand, all three changes offered by counterfactual examples contain variables that are consistent with the results of LIME and SHAP, showing the most important features.

PDPs can inherently only give a global interpretation. Looking at the patterns in PDP plots, one may understand how a variable affects the model's output, globally. Figure 3-c, for example, indicates that a higher heart rate decreases the likelihood of a negative result. This pattern is consistent with the global interpretation findings of SHAP and LIME; for example, samples with higher heart rates had lower SHAP values, contributing to a favorable outcome. PDPs can also be used to infer the significance of variables. For example, the slope of the plot for sodium is slightly greater than the slope for HR, indicating that sodium is more important. Because SHAP rates sodium higher than heart rate, this result is closer to SHAP than LIME. PDPs may depict the influence of no more than two features on the prediction result in a single plot. The contour of sodium and $SpO_2$ supports global interpretations by SHAP and LIME, demonstrating that $SpO_2$ has a modest influence on the outcome when compared to sodium. The key assumption in extracting PDPs is that the characteristics are independent. Consequently, PDPs are not applicable to time-series data because distinct time steps in a sequence (particularly neighboring time steps) are highly dependent.

### 4.2 Stability and Ease of Configuration

Based on our findings, the Anchors technique is heavily influenced by its hyperparameters. Even if the hyperparameters are specified, the outcomes of the approach may vary across runs. Similar to Anchors, the counterfactual example method includes a tolerance parameter, which specifies how close the counterfactual example predictions should be to the desired label. The resultant counterexample changes dramatically when the tolerance parameter is changed.

The number of characteristics on which the linear regression will be created must be specified by the user in LIME. We did not restrict the number of features, enabling the approach to estimate samples using the whole feature space. This might be one of the reasons why the LIME importance scores are more evenly distributed. SHAP, PDP, and explanation-by-examples approaches, on the other hand, do not need the user to tweak any major hyperparameters, and we found them to be the most stable.

### 4.3 Fidelity

SHAP had the highest fidelity likely because of its local accuracy property [16]. We showed that the SHAP contribution scores add up to the precise value of the prediction model output. LIME, on the other hand, does not have this property, as seen in Figure 2-b, where the LIME importance scores add up to 0.44, which is not the same as the prediction model output for patient A, i.e., 0.6. Anchors, on the other hand, provide high fidelity by default since they only provide rules with high precision, resulting in the method's adherence to the prediction model. Regarding the explanation-by-example method, the difference between the output of the supplied examples and the original sample may be seen as a fidelity measure. In Figure 5, the prediction model's output for the original sample is quite similar to the predictions of the instances, showing a relatively high fidelity. Other approaches, such as counterfactual example, are difficult to be evaluated regarding their fidelity since they only provide an instance with a different label comparing to the original instance.

### 4.4 Understandability

The Anchors technique, in our opinion, is simpler to grasp due to its rule-based nature, even if the results of this method for the given instance are inconsistent with the other three interpretation methods. Despite all of the advantages of LIME and SHAP, they seldom explain, in a human-understandable manner, why a certain output is predicted by the model. In physiologic data, for example, it is evident that the final time steps are important for all attributes, but the explanation is unclear. Are the final time steps of heart rate, for example, relevant since their values are decreasing? Or are they significant only because the values remain high enough despite the declining trend? A possible strategy for improving the understandability of SHAP/LIME interpretations, particularly in the clinical domain, is to augment them with clinician knowledge, i.e., clinician-in-the-loop interpretation.

The outcomes of the explanation-by-example technique are very hard to extract any information from. It is difficult to grasp the connections between the examples presented for the original sample. For example, there is little resemblance across the ICP plots (see Figure 5), which are the most essential characteristic for forecasting an unfavorable outcome. The original sample ICPs are about 21 with a high peak in the center, whereas one of the examples is around 20 with a peak at the end while the other one swings considerably, ranging from 15 to 30. As a result, it is hard for users to understand why and how the ICP values in the original sample contributed to the unfavorable outcome.

In general, example-based approaches function best when the feature values of an instance convey more context, such as images or texts. It is more difficult to describe tabular data in a meaningful fashion, because an instance might consist of hundreds or thousands of (less organized) characteristics. In this case, listing all feature values to describe an instance is ineffective. This was one of our challenges in showing the outcomes of counterfactual examples in the structured data. As a result, we decided to report only variables with more than a 0.5 change in their normalized value.

Since PDPs are in the form of plots, they are simple to comprehend. However, they do not transmit much information since one can only determine the influence of the variables on the prediction model's result. Their utility is likewise restricted since a partial dependency plot can only include two characteristics. This is not necessarily a limitation of PDPs, but rather our inability to easily envision more than three dimensions.

In conclusion, we suggest that SHAP is the most appropriate approach based on different XAI evaluation context. However, SHAP lacks understandability regarding both structured and time-series data. Table 2 summarizes our findings for recommended approaches across various XAI evaluation contexts.

**Table 2**. Recommended explanation methods for structured and physiologic data based on different XAI evaluation contexts

| Data | Fidelity | Stability and ease of configuration | Understandability |
|---|---|---|---|
| Structured Clinical Data | SHAP/Anchors | SHAP | Anchors |
| Physiological Data | SHAP | SHAP | SHAP/LIME |

## 5 Conclusion

We provided a comparison of six XAI approaches in the context of modeling clinical data. Our findings indicate that at the local level, SHAP and LIME results are extremely similar. Although its result is difficult to extrapolate information from, SHAP is the most stable with the highest fidelity and least configurable approach. Anchors, on the other hand, are the most understandable technique despite the fact that their results are not consistent with other approaches. Explanation-by-example, in our view, provides the less information about the mechanics of the models, making them the least understandable approach.

## 6 Acknowledgment

This material is based upon work supported by the National Science Foundation under grants #1838730 and #1838745. Dr. Foreman was supported by the National Institute of Neurological Disorders and Stroke of the National Institutes of Health (K23NS101123). The content is solely the responsibility of the authors. Any opinions, findings, and conclusions or recommendations expressed in this material are those of the authors and do not necessarily reflect the views of the National Science Foundation or of the National Institutes of Health. The authors acknowledge the TRACK-TBI Study Investigators for providing access to data used in this work.